\documentclass[a4paper, conference]{IEEEtran}
\usepackage{cite}
\usepackage{amsmath,amssymb,amsfonts}
\usepackage{algorithmic}
\usepackage{graphicx}
\usepackage{textcomp}
\usepackage{xcolor}
\usepackage{amsthm}
\usepackage{url}

\usepackage{comment}

\DeclareMathOperator{\silu}{SiLU}
\DeclareMathOperator{\softmax}{softmax}
\DeclareMathOperator{\ffn}{FFN}
\DeclareMathOperator{\LN}{LN}
\DeclareMathOperator{\localattn}{LocalSelfAttn}
\DeclareMathOperator{\cat}{Concat}
\DeclareMathOperator{\E}{E}
\DeclareMathOperator{\entropy}{H}
\newtheorem{theorem}{Theorem}

\newcommand{\kldiv}[2]{D_{KL}(#1\,\|\,#2)}
\newcommand{\stat}[3]{#1$\pm$#2(#3)}

\def\BibTeX{{\rm B\kern-.05em{\sc i\kern-.025em b}\kern-.08em
    T\kern-.1667em\lower.7ex\hbox{E}\kern-.125emX}}
\begin{document}

\title{Transformers discover an elementary calculation system exploiting local attention and grid-like problem representation}

\author{
\IEEEauthorblockN{Samuel Cognolato}
\IEEEauthorblockA{
\textit{University of Padova}\\
Padova, Italy\\
samuel.cognolato@studenti.unipd.it}
\and
\IEEEauthorblockN{Alberto Testolin}
\IEEEauthorblockA{
\textit{University of Padova}\\
Padova, Italy\\
alberto.testolin@unipd.it}
}

\maketitle

\begin{abstract}
Mathematical reasoning is one of the most impressive achievements of human intellect but remains a formidable challenge for artificial intelligence systems. In this work we explore whether modern deep learning architectures can learn to solve a symbolic addition task by discovering effective arithmetic procedures. Although the problem might seem trivial at first glance, generalizing arithmetic knowledge to operations involving a higher number of terms, possibly composed by longer sequences of digits, has proven extremely challenging for neural networks. Here we show that universal transformers equipped with local attention and adaptive halting mechanisms can learn to exploit an external, grid-like memory to carry out multi-digit addition. The proposed model achieves remarkable accuracy even when tested with problems requiring extrapolation outside the training distribution; most notably, it does so by discovering human-like calculation strategies such as place value alignment.
\end{abstract}


\section{Introduction}
Advanced mathematics is rooted in the acquisition of elementary concepts, such as number symbols and arithmetic operators. However, despite its apparent simplicity, learning to manipulate symbolic numbers is a sophisticated process that occupies children for several years during development and formal education \cite{ontogenetic_origins, number_sense}. Indeed, even mastering a basic procedure such as multi-digit addition involves a series of non-trivial skills: operands must be correctly aligned by place value, summations must be carried out in the proper order and regrouping must be performed by keeping track of the corresponding carry. Most importantly, the addition procedure should work for any number of operands, of any length.

The recent achievements of Artificial Intelligence (AI) in solving high-level reasoning tasks \cite{external_memory, rel_inductive_bias} have spurred interest in numerical cognition as a stimulating challenge for deep learning models \cite{dl_for_symb_math, math_reasoning, math_concepts}. Promising results have been obtained in a variety of domains, ranging from numerical reasoning over textual input \cite{math_language} to solving differential equations \cite{diff_equations} and automated theorem proving \cite{prove_theo_language}. However, deep learning often fails in elementary tasks that require systematic generalization: a prominent example is given by symbolic arithmetic, where neural networks do not easily extrapolate outside the numerical range encountered during training \cite{nalu}. Considering that digital calculators can solve such tasks in the blink of an eye, why is it so difficult to teach them to machine learning models? In trying to answer this question, we should keep in mind that it took centuries for humans to grasp even the most basic arithmetic principles, which were later implemented in digital calculators. Building machines that can autonomously discover algorithmic procedures might thus lay the foundations for creating more human-like artificial general intelligence.

In this paper we describe an innovative deep learning architecture that learns to generalize arithmetic knowledge well-beyond the numerical examples included in the training distribution. The model is trained on a set of multi-digit addition problems consisting of up to 4 operands, each composed by up to 10 digits; it is then tested over a much wider range of problems, featuring up to 10 operands and thousands of digits. The performance of the model is benchmarked against other recent models, and its internal functioning is investigated through ablation studies and analysis of the emerging internal representations.

\section{Related Work}
Most of contemporary machine learning approaches tackle symbolic arithmetic tasks by introducing explicit biases or human-engineered features specifically built for numerical reasoning. For example, the generalization performance of recurrent neural networks on single-digit addition was improved by designing activation functions enriched with primitive arithmetic operators \cite{nalu}, and further refinements of the same idea led to even higher extrapolation performance \cite{nau}. An alternative path is given by models that exploit an external memory to learn algorithmic tasks, such as Differentiable Neural Computers \cite{external_memory}, Grid LSTMs \cite{grid_lstm}, and Neural GPUs \cite{ngpu}. The latter two have been tested on multi-digit addition and multiplication, though generalization outside the training range was not systematically investigated for multi-operand problems (e.g., Neural GPU training examples included up to 20 bits and generalization was tested on problems of up to 2000 bits, but only for 2-terms additions).

One key property of algorithmic tasks is given by their sequential nature, which motivates the use of recurrent models. A particularly relevant architecture in this respect is the Universal Transformer \cite{universal_transformer}, which combines the parallelizability of feed-forward attention mechanisms with the inductive bias of recurrent networks. Being a parallel-in-time architecture, the Universal Transformer receives the entire series of input tokens at once; however, its recurrent nature allows to iteratively refine its internal state and thus produce output responses dynamically. Though such architecture was shown able to successfully learn a variety of algorithmic tasks, performance on integer addition was not satisfactory \cite{universal_transformer}.

Another important aspect to consider while learning an algorithmic task is that recurrent models should learn to run the necessary number of computational steps to process input sequences of different complexity. This problem can be tackled by embedding halting units into the model architecture, as in adaptive computation time (ACT) \cite{act} and PonderNet \cite{pondernet}.

Finally, it is well-known that certain arithmetic tasks (and multi-digit addition in particular) can be performed much easier and faster once numbers are aligned by place value. In agreement with this intuition, it has been shown that operand alignment indeed plays a key role to successfully learn symbolic addition with Neural GPUs \cite{ngpu_dec}. This finding motivated the design of more advanced mechanisms for input pre-processing, which allow to map the token sequence into a grid-like format to facilitate successive manipulation \cite{seq2grid}.

In this work we will combine several of the architectures and processing mechanisms reviewed above, with the goal of producing a comprehensive model that could more effectively tackle extrapolation in symbolic addition tasks. In Section \ref{sec:prop_appr} we will provide the formal details of our model, while in Section \ref{sec:exp_setup} we will describe the datasets, model parameters and training/testing details. Results and analyses will be presented in Section \ref{sec:results} and critically discussed in Section \ref{sec:conclusion}.

\section{Proposed Approach}
\label{sec:prop_appr}

\subsection{Problem Definition}
\label{subsec:prob_def}
The learning task considered in this work requires to sum an arbitrary number of operands, each composed of an arbitrary number of digits. These two degrees of freedom will be the main focus for measuring the extrapolation capabilities of the proposed architecture. An instance of the addition problem will be denoted by $\Pi([N_1,N_2],[D_1,D_2])$, where the four positive integers $[N_1,N_2]$ and $[D_1,D_2]$ denote the intervals for the number of operands and digits, respectively. For example, existing models such as the Grid LSTM \cite{grid_lstm} have been successful with $\Pi([2,2],[15,15])$, that is, sums of 2 operands of 15 digits each. The input for such addition problems can be devised as a symbol sequence $I\in\Sigma^S$, where $\Sigma=\{\text{PAD},0,1,\dots,8,9,+,=\}$ is the base-10 addition alphabet and $S$ is the length of the input sequence. $I$ is constrained to contain $n$ terms of $d_1,\dots,d_n$ digits, with $n\in[N_1,N_2]$ and $d_i\in[D_1,D_2]\;\forall i=1,\dots,n$.

Four properties should be taken into account when designing a model that can successfully solve this kind of task:
\begin{enumerate}
    \item Capability of manipulating discrete entities: humans break problems into easy-to-use parts that can be effectively manipulated and re-combined. Neural networks mimic this process when storing and moving data in an external memory \cite{external_memory} or when aggregating tokens in a sequence through self-attention mechanisms \cite{transformer}.
    \item Translation equivariance: a translation of the input should produce an equivalent translation of the output, which is useful for learning operators that do not depend on the absolute position they are applied to. In vision tasks this can be achieved by using convolutions \cite{shift_invariance} or relative positional encoding with self-attention \cite{transformer_tr_inv}.
    \item Permutation variance: permuting the order of the input could change the output. The majority of neural operators posses this property (e.g., permuting pixels in an image changes the activation of convolutional filters). On the contrary, self-attention without positional encoding is permutation invariant, as changing the order of tokens inside the sequence leads to the same output.
    \item Adaptive time: execution steps of an algebraic procedure should depend on the complexity of the input. In recurrent models the number of steps is often fixed \textit{a priori}: to adaptively stop execution when deemed fit, we must introduce dynamic halting mechanisms \cite{act, pondernet}.
\end{enumerate}

\subsection{Model Architecture}

\label{subsec:model_arch}
\begin{figure}[t]
    \centerline{\includegraphics[width=1.\linewidth]{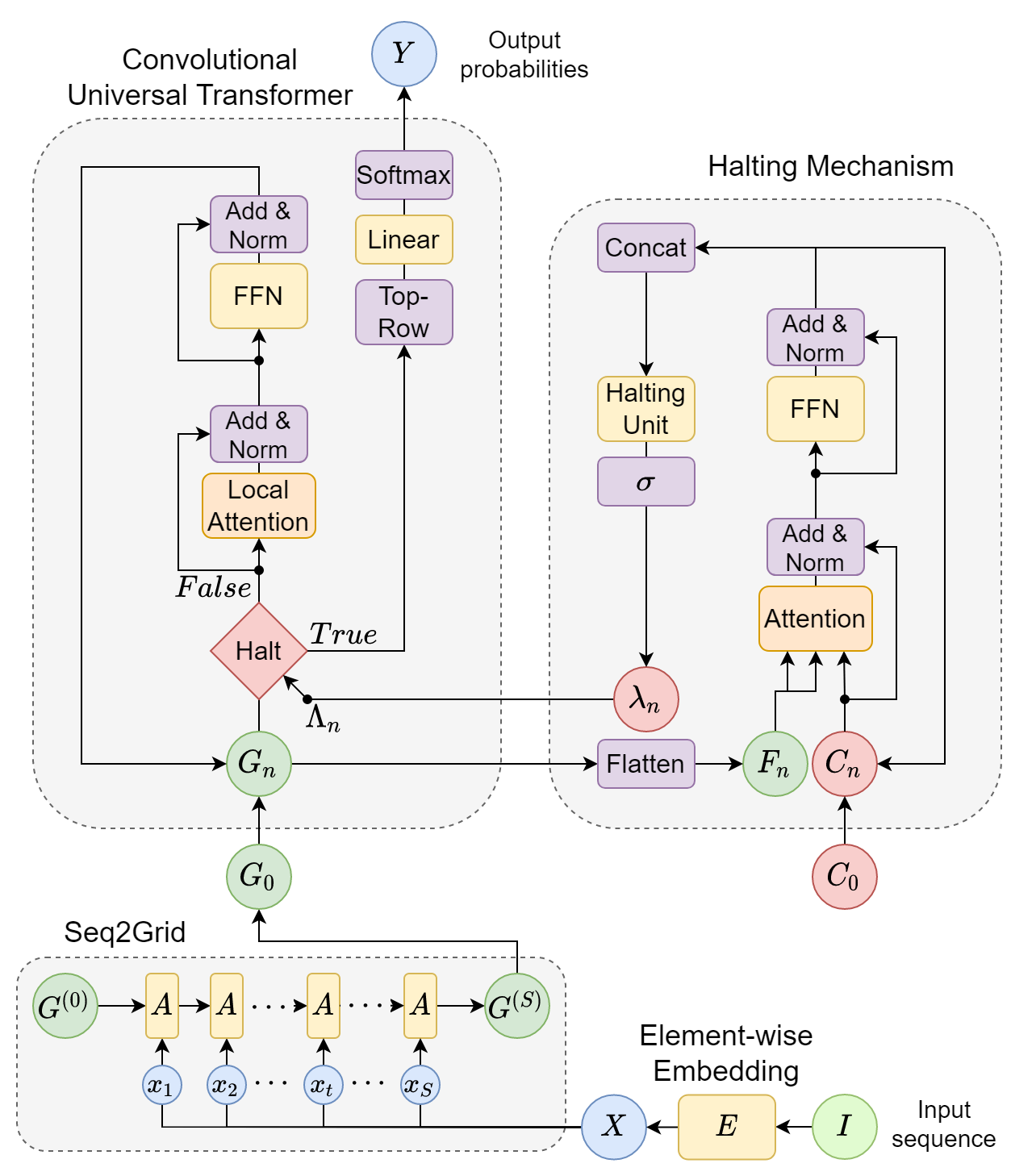}}
    \caption{High-level representation of the proposed architecture, which combines Seq2Grid preprocessing with a convolutional Universal Transformer with external memory and dynamic halting mechanisms.}
    \label{fig:full_architecture}
\end{figure}

Our architecture is built around the properties introduced in Sec. \ref{subsec:prob_def} and is composed of several modules working together (see Fig. \ref{fig:full_architecture}). For the scope of this paper, an $L$ layers feedforward network is defined as follows:
\begin{IEEEeqnarray}{c}
    \ffn_{L}(x)=
    \begin{cases}
        W_L \silu(\ffn_{L-1}(x)) + b_L & \text{if } L>1 \\
        W_1 x + b_1 & \text{if } L=1
    \end{cases} \label{eq:ffn}
\end{IEEEeqnarray}
where $W$ are the weights, $b$ are the biases and $\silu(\cdot)$ is the Sigmoid Linear Unit \cite{silu}. Processing is carried out through the following stages:

\subsubsection{Input and output}
\label{subsubsec:input_output}
The input symbol sequence $I\in\Sigma^S$ of length $S$, is first embedded into a corresponding vector sequence $X\in\mathbb{R}^{S\times d_{emb}}$ element-wise, using the learnable embedding matrix $E\in\mathbb{R}^{|\Sigma|\times d_{emb}}$, which maps each symbol in $\Sigma$ to a vector $x\in\mathbb{R}^{d_{emb}}$ in a lookup table fashion. Through learning, an embedding vector $x$ encodes in real numbers the meaning of its associated symbol, with no positional information, as the latter is added by local attention (Sec. \ref{subsubsec:ut}). $d_{emb}$ is the size of the embedding vectors, and is used throughout the whole architecture to comply with the recurrence of the architecture. As output, through a linear projection followed by the $\softmax$ function, the model produces a sequence of probabilities $Y$ of length $T$, such that each element is a distribution on the symbols of $\Sigma$. The output symbols can be picked as those with maximum probability.

\subsubsection{Seq2Grid Preprocessing}
\label{subsubsec:preproc}
The vector sequence $X$ is first preprocessed by rearranging the input vectors into a grid $G\in\mathbb{R}^{H\times W\times d_{emb}}$, where $H$ and $W$ are the fixed height and width of the grid. This enables the model to exploit useful structure in the input sequence that might not be evident in its 1-dimensional form. This stage is implemented using a Seq2Grid module \cite{seq2grid}, where grid operations are mirrored horizontally to make the grid readable and already in the right order for producing the output result. Input vectors are elaborated one at a time, choosing among three possible actions: insert the vector on the top row of the grid, shifting left all elements in that same row (Top List Update); insert the vector on a new empty row, shifting all elements down (New List Push); ignore the vector and hold the grid (No-Op). For each vector $x_t$, the action probabilities $a_{TLU}^{(t)}$, $a_{NLP}^{(t)}$, $a_{NOP}^{(t)}$ are computed through an encoder map, which in the original paper is a recurrent network. We opted for a simpler feedforward network, since in our case the rearrangement does not require to consider temporal dependencies:
\begin{equation}
    (a_{TLU}^{(t)}, a_{NLP}^{(t)}, a_{NOP}^{(t)})=\softmax(\ffn_2^{s2g}(x_t))
    \label{eq:seq2grid_controller}
\end{equation}
where the 2-layers $\ffn_2^{s2g}$ has hidden layers of size $d_{s2g}$. The initial grid $G^{(0)}$ is filled with zeroes. The intermediate grids $G^{(t)}$ $1\leq t\leq S$ are computed as:
\begin{IEEEeqnarray*}{rl}
    G^{(t)}=\;&a^{(t)}_{TLU}TLU(G^{(t-1)},x_t) +\\
    &+a^{(t)}_{NLP}NLP(G^{(t-1)},x_t)+a^{(t)}_{NOP}G^{(t-1)}\\
    TLU&(G,x)_{i,j}=
    \begin{cases}
        x & \text{if }i=1,j=W\\
        G_{1,j+1} & \text{if }i=1,j<W\\
        G_{i,j} & \text{if }i>1\\
    \end{cases} \yesnumber\\
    NLP&(G,x)_{i,j}=
    \begin{cases}
        x & \text{if }i=1,j=W\\
        0 & \text{if }i=1,j<W\\
        G_{i-1,j} & \text{if }i>1\\
    \end{cases}
    \label{eq:seq2grid_mirrored}
\end{IEEEeqnarray*}

\subsubsection{Universal Transformer with Local Attention}
\label{subsubsec:ut}
The resulting grid $G^{(S)}$ will now be denoted as $G_0$, as it undergoes several computational steps through a Universal Transformer \cite{universal_transformer}. The number of computational steps is decided by a 2-state stochastic process $\Lambda_n$ where state 0 means ``continue'' and state 1 means ``stop'' (the halting mechanism is described in detail below). After halting, the output of the network is directly read from the top row of the grid. A single computational step is defined as:
\begin{equation}
    G_{n+1}=\text{UT}(G_n)=
    \begin{cases}
        \text{ConvTransfBlock}(G_n) & \text{if }\Lambda_n=0 \\
        G_n & \text{if }\Lambda_n=1
    \end{cases}
    \label{eq:main_ut}
\end{equation}
where the ConvTransfBlock is a convolutional transformer block implementing the core of computation on the grid $G_n$. It is implemented as a standard transformer block \cite{transformer} with a local self-attention:
\begin{equation}
    \label{eq:transformer_block}
    \begin{split}
        G'&=\LN(X_n+\localattn(G_n)) \\
        G_{n+1}&=\LN(G'+\ffn_3^{ut}(G'))
    \end{split}
\end{equation}
where $\LN$ is the Layer Normalization and the $\localattn$ operator is an extension of the Stand-Alone Self Attention (SASA) \cite{sasa} with number of groups $g$ and number of heads $h$. The vectors in the input grid $G\in\mathbb{R}^{H\times W\times d_{emb}}$ are first split into $g$ groups $G^l\in\mathbb{R}^{H\times W\times d_{emb}/g}$, then separately and linearly projected into queries, keys, and values:
\begin{equation}
    Q^l=G^lW_Q^l, \; K^l=G^lW_K^l, \; V^l=G^lW_V^l, \quad 1\leq l\leq g
\end{equation}
where $W_Q^l,W_K^l,W_V^l\in\mathbb{R}^{d_{emb}/g\times d_{emb}/g}$ are weight matrices. Queries, keys and values are concatenated and split again into $h$ parts $Q^m,K^m,V^m$, one for each head. Values are then aggregated through a convolutional operator with weights computed from the usual dot-product:
\begin{IEEEeqnarray*}{c}
    Y_{ij}^m=\sum_{a,b\in\mathcal{N}_k(i,j)}A_{ij,ab}^m V_{ab}^m \\
    A_{ij,ab}^m=\softmax_{ab}((Q_{ij}^m+s^m)^T(K_{ab}^m+r_{a-i,b-j}^m)) \yesnumber
\end{IEEEeqnarray*}
where $A_{ij,ab}^m$ is the attention that position $ij$ pays to position $ab$ at head $m$, $r_{a-i,b-j}\in\mathbb{R}^{k\times k\times d_{emb}}$ is a learned relative positional encoding, $s\in\mathbb{R}^{d_{emb}}$ is a learned query encoding, $\mathcal{N}_k(i,j)$ is the neighborhood of position $ij$ with spatial extent $k$. Our definition differs from \cite{sasa} in 2 points: (1) we split vectors at two different points, contrary to SASA which can be interpreted as the special case $g=h$; (2) we added the query encoding $s$ and extended $r_{a-i,b-j}$. The use of both $r_{a-i,b-j}$ and $s$ is meant to allow the network to build more expressive rules for aggregating value vectors. This can be seen by expanding the expression:
\begin{IEEEeqnarray}{rCl}
    \IEEEeqnarraymulticol{3}{l}{
    (Q_{ij}+s)^T(K_{ab}+r_{a-i,b-j})=
    }\nonumber\\* \quad
    & = & Q_{ij}^TK_{ab}+Q_{ij}^Tr_{a-i,b-j}+s^TK_{ab}+s^Tr_{a-i,b-j} \label{eq:attn_rules_terms}
\end{IEEEeqnarray}
The model is free to learn rules where some features are aggregated independently from queries and keys, but only based on the relative position information contained in $s^Tr_{a-i,b-j}$ when this term dominates the sum. Likewise, positional information can be partially or totally ignored when the opposite happens for other content-based terms. In other words, the attention payed to tokens can depend on the content of tokens (content-based rules), on the position (position-based rules), or a mixture of the two. Empirical analyses presented later show that the model in fact learns all these kind of rules (Sec. \ref{subsec:computation}).

\subsubsection{Halting Mechanism}
\label{subsubsec:halting_mech}
As halting policies we consider both a fixed number of steps and a more sophisticated PonderNet policy \cite{pondernet}. In the latter, a separate network produces a single conditioned halting probability at each time step $n$:
\begin{equation}
    \lambda_n=Pr\{\Lambda_n=1|\Lambda_{n-1}=0\}\quad 0\leq n\leq N
\end{equation}
with the Markov process $\Lambda_n$ starting at $\Lambda_{-1}=0$. The \textit{a priori} probability distribution $p_n$ can be computed as a truncated generalized geometric distribution:
\begin{equation}
    p_n=\lambda_n\prod_{i=0}^{n-1}(1-\lambda_i),\quad p_N=1-\sum_{i=0}^{N-1}p_i \\
\end{equation}
where $N$ is the minimum number of steps at which the cumulative distribution exceeds the threshold $1-\epsilon$, where $\epsilon$ is a small hyperparameter. During training, the expected value of all losses computed at each time step is taken with probability distribution $p_n$, unlike ACT where only one loss is computed from the expected value of all outputs. This is a significant difference, as in the former the output result does not depend on the distribution $p_n$ if not for stopping, whereas the latter takes weighted sums of its internal values, also at evaluation time. PonderNet simplifies halting at evaluation time, because halting events can be sampled as a Bernoulli of probability $\lambda_n$ ($\Lambda_n\sim B(\lambda_n)$). The full PonderNet loss is:
\begin{equation}
    \label{eq:objective_pondernet}
    \hat{\mathcal{L}}(y,\hat{y}_n)=\sum_{n=0}^N p_n\mathcal{L}(y,\hat{y}_n)+\beta R(p_n)
\end{equation}
where $y$ is the ground truth and $R(p_n)$ is a regularizer for the \textit{a priori} distribution $p_n$, weighted by the hyperparameter $\beta$. The original paper uses $R(p_n)=\kldiv{p_n}{p_G(\lambda_p)}$ in order to regularize $p_n$ as a geometric distribution $p_G$ (truncated at $N$) of parameter $\lambda_p$. According to the authors, this incentivizes exploration giving a nonzero probability to all possible steps, while the model learns to use computational time efficiently as a form of Occam's Razor.

By extending the definition of KL-divergence with a geometric distribution, it is possible to extract its dependence from the negative entropy $-H(p_n)$ and expected number of steps before halting $E[N_h]$ where $N_h$ is distributed as $p_n$. Motivated by these considerations, we propose a new hyperparameter-free regularizer for $p_n$ named Explore-Reinforce (ER), which contains a reformulation of the two terms mentioned above:
\begin{equation}
    R_{ER}(p_n, a)=\underbrace{-(1-a)\entropy(p_n)}_{\text{Explore}}+\underbrace{a\E[log(1+N_h)]}_{\text{Reinforce}}
\end{equation}
where $a\in[0,1]$ is a sample-wise measure of success, such as the model per-sample accuracy. It is trivial to see that both the entropy and expected log value are constrained in the interval $[0,log(1+N)]$, and the hyperparameter $\lambda_p$ is dropped. The accuracy trades-off the Explore and Reinforce terms: while the model learns to solve the problem, it is incentivized to give equal possibilities to each step by maximizing the entropy of $p_n$; as the model progresses, computation steps have to be compressed and made more efficient by minimizing the expected log-number of steps. In other words, the model learns to take fewer steps for easier problems.

\subsubsection{Halting through a Context Transformer}
\label{subsubsec:context_transf}
As elegant as the PonderNet formulation is, it is not obvious how to fit it in the definition of the Universal Transformer, which splits the halting process to the token level using ACT. Instead, we pair an halting transformer to the main Convolutional UT, gathering information from the grid into a compact context sequence $C_n\in\mathbb{R}^{S_C\times d_{emb}}$, where $S_C$ is the length of the sequence, fixed as hyperparameter. Each element of the context sequence can attend to the grid $G_n\in\mathbb{R}^{H\times W\times d_{emb}}$ through an attention mechanism by flattening the 2D grid into a 1D sequence $F_n\in\mathbb{R}^{HW\times d_{emb}}$. An additional row-wise position encoding, defined as an adaptation of the simple ALiBi encoding \cite{alibi}, is added inside the $\softmax$ operator of the attention mechanism:
\begin{equation}
\label{eq:alibi}
    \text{ALiBi}(Q,K,V)=\softmax\bigg(\frac{QK^T}{\sqrt{d_k}} + m\cdot M\bigg)V \\
\end{equation}
where $m$ is an head-specific slope and $M$ assigns decreasing scores, as defined in the original ALiBi paper \cite{alibi}. $M$ is adapted to a grid setup by assigning the same score for each element in the same row, starting at $0$ for the top row, $-1$ for the one below and so on. It has been shown that ALiBi can reduce training time, increase generalization, and avoid hyperparameters in training transformers \cite{alibi}. Having defined the attention mechanism, the halting transformer is thus built as follows:
\begin{equation}
\label{eq:context_transformer}
\begin{split}
    C_n'&=\LN(C_n+\text{ALiBi}(C_n, F_n, F_n)) \\
    C_{n+1}&=\LN(C_n'+FFN^{ctx}_2(C_n')) \\
    \lambda_n&=\sigma(FFN^{halt}_2(\cat(C_{n+1})))
\end{split}
\end{equation}
where the initial context sequence $C_0\in\mathbb{R}^{S_C\times d_{emb}}$ is initialized as a learnable weight matrix, and $\lambda_n$ is the conditioned halting probability from Sec. \ref{subsubsec:halting_mech}.

\section{Experimental setup}
\label{sec:exp_setup}
\subsection{Datasets}
Several configurations of the model architecture described above are trained on the problem instance $\Pi([1,4],[1,10])$, i.e., additions of 1 to 4 terms, each made of 1 to 10 digits. For each training example, a uniform random number in $[1,4]$ is picked as \#terms, and for each term a uniform random number in $[1,10]$ is picked as its \#digits, where each digit is uniformly picked from $\{0,1,\dots,9\}$. This allows to have a fair distribution of examples with respect to sequence length\footnote{Sampling terms directly from $[0,10^{11}-1]$ would bias the distribution towards higher numbers, as there would be a $9/10$ probability of generating $10$-digits numbers, but only $9/100$ for $9$-digits numbers, and so on.}. The $+$ symbol links each sampled term, and the $=$ symbol terminates the string. PAD symbols are appended to equalize lengths and allow to group sequences in batches. As output target, the model only receives the correct result of the sum.

Generalization performance is tested on problem instances that have been solved to perfection ($>$99\% sequence accuracy) by related architectures: additions of 2 numbers of 15 digits each, solved by 2-LSTM \cite{grid_lstm}; additions of 2 numbers of 100 digits, solved by the Neural GPU \cite{ngpu_dec}; additions of 2 numbers of 602 digits, coarsely corresponding to the maximum length of 2000 binary digits also used to test the Neural GPU \cite{ngpu}; additions of 1 to 5 numbers of 1 to 5 digits, solved by an LSTM using ACT \cite{act}. It should be noted that the Neural GPU required a carefully tuned curriculum learning to reach this level of performance, and the LSTM+ACT model needed supervision on intermediate results to solve additions featuring many operands. We also include further test cases in order to better explore the extrapolation capability on the number of terms.

\subsection{Model parameters and architectural variants}
\label{subsec:model_params}
The main parameters of our base model are summarized in Table \ref{table:model_params}\footnote{Note that the embedding size $d_{emb}$ is set as the tokens vector size throughout the whole network, from input embeddings to output sequence, including the context vectors. Dropout is used after all linear layers.}. To better investigate the role of each processing module we also test five different model variants, reported in Table \ref{table:all_models}, disabling or changing single components one at a time. Our base model has 1 local attention head for each element of the embedding vector; noGroups resembles the original self-attention definition \cite{transformer} regarding linear projections and number of heads; the SASA variant follows exactly the definition in \cite{sasa}; fixedTime uses a constant number of recurrent steps (set to $12$), without any dynamic halting mechanism; and PonderReg uses the usual PonderNet regularization, that is, the KL-divergence from a geometric distribution.
\begin{table}[t]
    \centering
    \caption{Model parameters.}
    \begin{tabular}{|c | c|} 
        \hline
        \textbf{Parameter name} & \textbf{Value}\\
        \hline
        Embedding size $d_{emb}$            & $64$ \\
        Dropout $p$                         & $0.1$ \\
        \hline
        Internal dimension of $FFN_2^{s2g}$ & $64$ \\
        \hline
        Spatial extent $k$ (kernel size)    & $3$\\
        Local attention groups $g$          & $8$ \\
        Local attention heads $h$           & $64$ \\
        Internal dimension of $FFN_3^{ut}$  & $256$ \\
        \hline
        Internal dimension of $FFN_2^{ctx}$     & $64$ \\
        Internal dimension of $FFN_2^{halt}$    & $128$ \\
        Context length $S_C$                    & $3$ \\
        Context attention heads                 & $8$ \\
        Maximum number of steps                 & $40$ \\
        Distribution's threshold $\epsilon$     & $0.05$ \\
        \hline
    \end{tabular}
    \label{table:model_params}
\end{table}

\begin{table}[t]
    \centering
    \caption{Model variants.}
    \begin{tabular}{l c c c c c}
        \hline
        \textbf{Model}  & $g$ & $h$ & dynamic halting & $p_n$ reg.\\
        \hline
        base            & 8 & 64 & \checkmark & ER \\
        noGroups ($g=1$)& 1 & 64 & \checkmark & ER \\
        SASA ($g=h$)    & 8 & 8  & \checkmark & ER \\
        fixedTime       & 8 & 64 &            &  \\
        ponderReg       & 8 & 64 & \checkmark & KL-div \\
        \hline
    \end{tabular}
    \label{table:all_models}
\end{table}

The grid sizes $H,W$ (Sec. \ref{subsubsec:preproc}) are always shared in a single batch of examples during training. Let $N$ and $D$ be the batch maximum number of terms and digits respectively. Then we set $H,W$ as:
\begin{IEEEeqnarray}{c}
    H=N+F_H+R_H \\
    W=D+F_W+R_W
\end{IEEEeqnarray}
where $F_H$, $F_W$ are two fixed scalars, $R_H$, $R_W$ are two uniform random variables. $F_H$ and $F_W$ are ``oversizes'' that leave the model space to work and eventually produce longer outputs. $R_H$ and $R_W$ are ``regularizers'' that help avoiding overfitting on a fixed grid size. To lower the computational cost of the model, we chose some small values: $F_H=0, F_W=2, R_H\in[0,3], R_W\in[0,3]$, and divided samples from the dataset into groups based on the number of terms in order to lower $H,W$ per batch. This choice won't affect training as the gradients are computed on all losses from each group reduced together. We chose two groups of $[1,2]$ and $[3,4]$ number of terms, and same $[1,10]$ number of digits.

\subsection{Training details}
\label{subsec:training}
We adopt a cross entropy loss, where the PAD class is weighted 1/10 of other classes to balance its higher frequency of appearance and PonderNet regularizers are weighted by $\beta=$5e-2. All models are trained for 510 epochs of 10 training steps each, using the AdamW optimizer \cite{adamw} with the standard parameters $\beta_1=0.9$, $\beta_2=0.999$ and a weight decay of $0.1$ for all weights, excluding biases and embeddings. We employ a cosine annealing schedule with a learning rate $\eta$ ranging from 1e-3 to 5e-5 over a period of 30 epochs. Gradient norm is clipped to 10 to avoid exploding gradients. A batch size of 128 is used, making 64 samples for each of the two groups defined in Sec. \ref{subsec:model_params}. For each variant, we train 10 models initialized with different random seeds. The best models in terms of extrapolation of \#digits and \#terms are then selected and further trained for other 300 epochs in an ``overtraining'' phase, lowering the PonderNet weighting to $\beta=$5e-4 as the regularizer dominated the loss during the last epochs of the standard training phase. All models are trained using an NVIDIA Tesla K80 GPU\footnote{PyTorch source code available at \url{https://github.com/CognacS/tag-cat}}.

\begin{table*}[t]
    \caption{Accuracy on different problem instances requiring extrapolation on the number of digits and number of operands.}
    \centering
    \begin{tabular}{|l | cc cc cc | cc cc cc cc |}
        \hline
         & \multicolumn{6}{ c|}{2-terms additions} & \multicolumn{8}{ c|}{N-terms additions} \\
        \textbf{Model} & \multicolumn{2}{ c }{$15$ digits} & \multicolumn{2}{ c }{$100$ digits} & \multicolumn{2}{ c|}{$602$ digits} & \multicolumn{2}{ c }{$[1,4],[1,10]$} & \multicolumn{2}{ c }{$[1,5],[1,5]$} & \multicolumn{2}{ c }{$[5,6],[1,10]$} & \multicolumn{2}{ c|}{$[7,10],[1,10]$}\\
        \cline{2-13}
        & char & seq & char & seq & char & seq & char & seq & char & seq & char & seq & char & seq\\
        \hline
        base model & \textbf{1.0} & \textbf{1.0} & \textbf{0.99} & \textbf{0.99} & 0.98 & 0.0 & \textbf{1.0} & \textbf{1.0} & \textbf{1.0} & \textbf{1.0} & \textbf{0.99} & \textbf{0.93} & \textbf{0.72} & \textbf{0.25}\\
        noGroups variant ($g=1$) & \textbf{1.0} & \textbf{1.0} & \textbf{0.99} & \textbf{0.99} & \textbf{0.99} & \textbf{0.99} & 0.99 & 0.99 & 0.99 & 0.99 & 0.98 & 0.86 & 0.54 & 0.1\\
        \hline
    \end{tabular}
    \label{table:comparison_results}
\end{table*}

\begin{table}[t]
    \centering
    \caption{Accuracy of the noGroups variant on additions featuring 2 terms composed by many digits (left). Accuracy of the base model on additions featuring $N$ terms of just 5 digits (right).}
    \begin{tabular}{c c c}
        \hline
        \#digits & char & seq\\
        \hline
        $[1\quad\;\;,1000]$      & 0.99 & 0.99 \\
        $[1000,1000]$            & 0.99 & 0.99 \\
        $[1\quad\;\;,2000]$      & 0.99 & 0.99 \\
        $[2000,2000]$            & 0.99 & 0.81 \\
        $[1\quad\;\;,4000]$      & 0.99 & 0.58 \\
        $[4000,4000]$            & 0.86 & 0.0 \\
        \hline
    \end{tabular}\quad\quad\quad
    \begin{tabular}{c c c}
        \hline
        \#terms & char & seq\\
        \hline
        $5$   & 1.0 & 1.0 \\
        $6$   & 0.99 & 0.98 \\
        $7$   & 0.98 & 0.86 \\
        $8$   & 0.84 & 0.40 \\
        $9$   & 0.66 & 0.09 \\
        $10$  & 0.48 & 0.03 \\
        \hline
    \end{tabular}
    \label{table:long_additions}
\end{table}

\begin{figure}[t]
    \centering
    \includegraphics[width=0.48\textwidth]{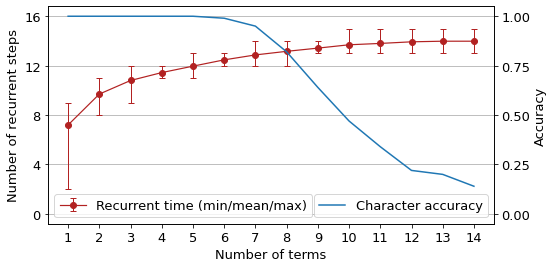}
    \caption{Correlation between the number of terms, recurrent steps before halting, and accuracy at character and sequence level when computing with our base overtrained model.}
    \label{fig:rec_steps_acc}
\end{figure}

\subsection{Evaluation Metrics}
\label{subsec:eval_metrics}
Task accuracy is computed by dividing the number of correct matches by the number of valid matches. As valid matches, we consider three scenarios: number-number, pad-number, number-pad, and ignore all correct pad-pad matches. This ensures that the accuracy measure is not inflated by the high frequency of correct pad-pad matches. Accuracy is computed both at \textit{character level}, that is, all characters for valid matches are considered, and at \textit{sequence level}, where any error in the sequence invalidates the entire sample.

\section{Results}
\label{sec:results}

\subsection{Generalization capabilities}
Results achieved by the two best overtrained models are reported in Table \ref{table:comparison_results} (note that accuracy values in the interval $[0.99, 1.0)$ are always rounded down to $0.99$, as we only consider $1.0$ to be a perfect score). Both models match the performance of state-of-the-art approaches, at the same time exhibiting remarkable accuracy on novel problem instances featuring more challenging extrapolation ranges over the number of operands (N-terms additions). Interestingly, the noGroups variant achieves better extrapolation on the number of digits, while the base model exhibits better extrapolation on the number of terms. In Table \ref{table:long_additions} we push these tests to the limit, showing that additions of 2 very long numbers can still be solved with high accuracy, while extrapolation on the number of terms appears more challenging.

Fig. \ref{fig:rec_steps_acc} suggests a possible correlation between the drop in accuracy and the number of recurrent steps before halting, which seems to stabilize even if the increasing number of terms might in fact require more computing steps. We also encountered the same problem discussed in \cite{ngpu_dec}: models often fail when trying to carry over lengths higher than those found during training. Some representative examples are shown in Table \ref{table:examples_errors}, where changing the order of terms surprisingly returns different results. Errors occur when a large operand is followed by smaller ones (which requires to move all digits over long distances), when the carry must be iteratively propagated, or when the number of terms exceeds a certain value.

\begin{table}[t]
    \caption{Handpicked representative examples.}
    \centering
    \begin{tabular}{r c c c}
        \hline
        Operation & Network pred. & True result & Correct\\
        \hline
        11134+1+1+1+1+1= & 139 & 11139 &\\
        1+1+1+1+1+11134= & 11139 & 11139 & \checkmark\\
        1+1+1+1+1+1+1+11134= & 11140 & 11141 &\\
        99999+1= & 100000 & 100000 & \checkmark \\
        999999999+1= & 999000000 & 1000000000 &\\
        999990000+9999+1= & 990000000 & 1000000000 &\\
        1+1+1+1+1+1= & 6 & 6 & \checkmark \\
        1+1+1+1+1+1+1= & 6 & 7 &\\
        \hline
    \end{tabular}
    \label{table:examples_errors}
\end{table}

\begin{figure}[t]
    \centering
    \includegraphics[width=0.47\textwidth]{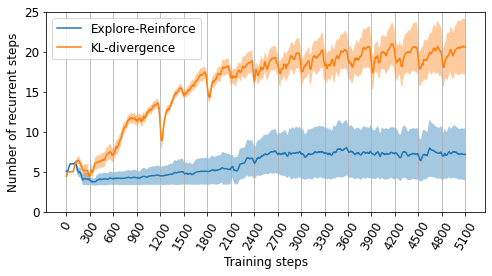}
    \caption{Mean number of recurrent steps at training time learned by regularizing the halting distribution with our Explore-Reinforce and the KL-divergence with a geometric distribution. Colored areas represent the standard deviation. Jumps in recurrent steps are caused by the cosine annealing schedule.}
    \label{fig:abl_rec_steps_reg}
\end{figure}

\begin{table*}[t]
    \caption{Statistics on accuracy scores of different variants of our proposed model. Each variant was trained with 10 different initializations and the average $A$, standard deviation $SD$ and max score $M$ are reported with the format $A\pm SD(M)$.}
    \centering
    
    \begin{tabular}{|l | cc cc cc |}
        \hline
        \textbf{Model} & \multicolumn{2}{ c }{$[2,2],[602,602]$} & \multicolumn{2}{ c }{$[5,6],[1,10]$} & \multicolumn{2}{ c|}{$[7,10],[1,10]$}\\
        \cline{2-7}
        & char & seq & char & seq & char & seq\\
        \hline
        base       & \stat{0.23}{0.27}{0.98} & \stat{0.00}{0.00}{0.00} & \stat{\textbf{0.93}}{0.09}{0.98} & \stat{\textbf{0.72}}{0.24}{\textbf{0.90}} & \stat{0.51}{0.14}{0.66} & \stat{\textbf{0.09}}{0.05}{0.16} \\
        noGroups   & \stat{\textbf{0.99}}{0.00}{\textbf{0.99}} & \stat{\textbf{0.63}}{0.34}{0.93} & \stat{0.74}{0.26}{0.96} & \stat{0.40}{0.27}{0.75} & \stat{0.29}{0.13}{0.51} & \stat{0.02}{0.02}{0.06} \\
        SASA       & \stat{0.70}{0.39}{\textbf{0.99}} & \stat{0.20}{0.33}{\textbf{0.95}} & \stat{0.81}{0.29}{0.98} & \stat{0.60}{0.30}{0.85} & \stat{0.45}{0.17}{0.64} & \stat{0.05}{0.04}{0.12} \\
        fixedTime  & \stat{0.44}{0.40}{\textbf{0.99}} & \stat{0.12}{0.25}{0.70} & \stat{0.76}{0.16}{0.95} & \stat{0.36}{0.27}{0.74} & \stat{0.27}{0.11}{0.42} & \stat{0.01}{0.01}{0.04} \\
        ponderReg  & \stat{0.19}{0.14}{0.52} & \stat{0.00}{0.00}{0.00} & \stat{\textbf{0.93}}{0.06}{\textbf{0.99}} & \stat{0.69}{0.21}{\textbf{0.90}} & \stat{\textbf{0.54}}{0.16}{\textbf{0.72}} & \stat{\textbf{0.09}}{0.08}{\textbf{0.21}} \\
        \hline
    \end{tabular}
    
    \label{table:ablation_results}
\end{table*}

\begin{figure*}[t]
    \centering
    \raisebox{3mm}{\includegraphics[width=0.68\textwidth]{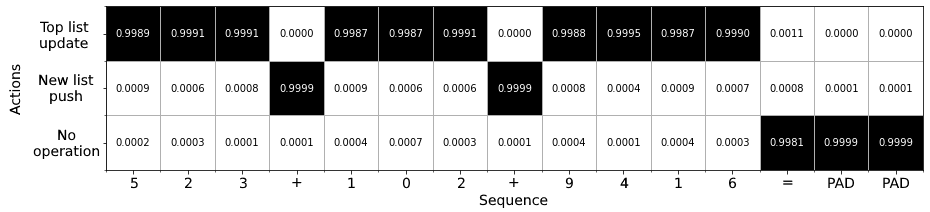}}
    \includegraphics[width=0.30\textwidth]{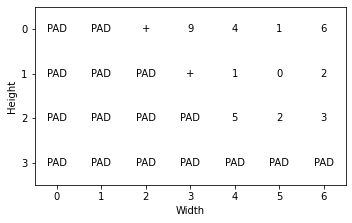}
    \caption{Action probabilities (left) and resulting grid $G^{(S)}=G_0$ from the padded input sequence ``523 + 102 + 9416 = $<$PAD$>$ $<$PAD$>$''.}
    \label{fig:vis_grid_making}
\end{figure*}

Table \ref{table:ablation_results} reports the average scores, standard deviation, and max scores of different variants of the base model. These results confirm that perfect extrapolation on both \#digits and \#terms never happens, and model components seem to specialize in tackling one of these two degrees of freedom. The noGroups variant is the most solid on the digits extrapolation task $\Pi([2,2],[602,602])$, solved consistently with $>0.99$ accuracy, while other methods achieve perfect accuracy only with some lucky initialization. On the tasks requiring extrapolation over the number of terms $\Pi([5,6],[1,10])$ and $\Pi([7,10],[1,10])$, the original SASA variant is outperformed by our base model, and fixing the number of recurrent steps further degrades performances. The accuracy scores of the base and ponderReg variants are comparable; however, it should be noted that training the latter took 4 hours, compared with the 2 hours required by our base approach. This phenomenon can be explained by comparing the halting steps during training, as shown in Fig. \ref{fig:abl_rec_steps_reg}: our Explore-Reinforce regularization allows to learn a much more efficient criterion for halting.

\begin{figure}[t]
    \centering
    \includegraphics[width=0.17\textwidth]{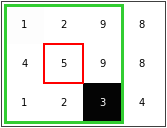}
    \includegraphics[width=0.17\textwidth]{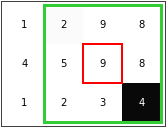}
    
    \vspace{3mm}
    
    \includegraphics[width=0.17\textwidth]{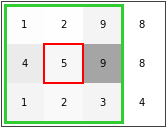}
    \includegraphics[width=0.17\textwidth]{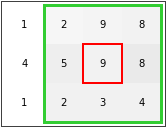}
    \caption{Examples of a position-based rule (top) and content-based rule (bottom).}
    \label{fig:attn_rules}
\end{figure}

\subsection{Analysis of step-by-step computation}
\label{subsec:computation}
In this section we explore how the model learned to solve the addition problem. Indeed, the use of sigmoid and $\softmax$ activations in focal points of the network, such as the Seq2Grid actions or the attention aggregation, increases its explainability, since importance is reflected on the magnitude of the activation.

\subsubsection{Seq2Grid} The preprocessing module learned a reliable procedure to format the incoming sequence into a grid. As shown in Fig. \ref{fig:vis_grid_making}, all digits are appended to the top list while + signs break the row and push a new line. Equals signs and paddings are correctly ignored as they do not contribute to the final evaluation. This formatting is actually the same we humans use when solving additions with the column method.

\begin{figure}[t]
    \centering
    \includegraphics[width=0.35\textwidth]{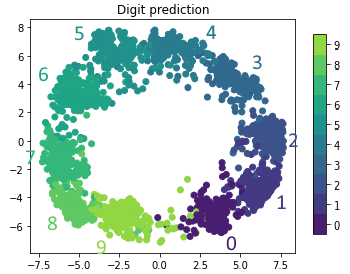}
    \caption{PCA projection of token vectors produced by the model, highlighting the emergence of a ring-shaped structure.}
    \label{fig:emerging_repr}
\end{figure}

\subsubsection{Local Attention} The transformer learned to produce different kinds of ``rules'' used by attention heads to aggregate neighboring tokens. We found all of the rules explained in Sec. \ref{subsubsec:ut}; in particular, most position-based rules (example in top row of Fig. \ref{fig:attn_rules}) are used to aggregate a single token in a specific neighborhood location. More complex content-based rules seem to depend on the magnitude of digits: the example in bottom row of Fig. \ref{fig:attn_rules} shows that attention can focus on large digits in the neighboring row positions, but can also be payed to all surroundings in the case of large querying digits.

\subsection{Emergent internal representations}
We finally investigated how token vectors are manipulated by the model by visualizing its representational space using Principal Component Analysis. To do so, we sampled a large batch of different problems and plotted the first two components of the vectors extracted at each time step, colored according to the corresponding symbol produced by the model  (plus, equal and PAD are ignored). As shown in Fig. \ref{fig:emerging_repr}, it is evident that the representational space self-organizes according to a ring-shaped structure, where vectors of the same class are clustered together and digits are ordered from 0 to 9, and then back to 0. Such structure makes sense, because it allows to linearly change the magnitude of produced digits by moving between adjacent clusters; moreover, when vectors corresponding to high valued digits have to propagate a carry they simply cycle back to zero, thus allowing to restart the incremental process.

\section{Discussion and Conclusion}
\label{sec:conclusion}
In this paper, we proposed a sophisticated yet lightweight deep learning model, assembling a variety of architectures and processing mechanisms with the aim of studying how neural networks could learn to solve multi-digit addition and generalize arithmetic knowledge to novel problems. The proposed model matches current state-of-the-art approaches on problems involving 2-operands that require extrapolation over the number of digits, at the same time exhibiting improved generalization on problems involving more operands. A distinguishing feature of our model is the use of a novel centralized halting mechanism, compatible with the definition of Universal Transformers and PonderNet, which allows to speed-up learning by calibrating the number of computational steps required to solve problems of different complexity.

It is well-known that the lack of explicit inductive biases makes it very challenging for neural networks to extrapolate well on arithmetic problems. In this respect, our simulations suggest that equipping deep learning agents with external memory systems might be a key principle to promote systematic abstraction when learning algorithmic tasks. At the same time, the capability of our model to extrapolate on problems with a large number of operands is still fairly limited, motivating further efforts to improve neural network models of mathematical symbol grounding \cite{symbol_grounding_prob_2}. For example, future research could investigate whether generalization performance might improve by grounding arithmetic procedures on perceptual representations of numbers \cite{number_sense_emergentist, number_sense_visual_magnitude} and/or more advanced external representations mimicking the calculation tools invented by human cultures \cite{self_com_drl_number_repr}.

\end{document}